\documentclass{article}

\usepackage[preprint,nonatbib]{neurips_2022}

\usepackage[utf8]{inputenc} 
\usepackage[T1]{fontenc}    
\usepackage{url}            
\usepackage{booktabs}       
\usepackage{amsfonts}       
\usepackage{nicefrac}       
\usepackage{microtype}      

\usepackage[ruled,vlined,linesnumbered]{algorithm2e}

\usepackage{preamble}

\title{Computing Abductive Explanations\\ for Boosted Trees}

\author{
Gilles Audemard$^1$\And
Jean-Marie Lagniez$^1$\And
Pierre Marquis$^{1, 2}$ 
\And
Nicolas Szczepanski$^1$\\
~\\
$^1$CRIL, Université d'Artois \& CNRS, France\\
$^2$Institut Universitaire de France, France\\
~\\
\{audemard, lagniez, marquis, szczepanski\}@cril.fr
}

\begin{document}

\maketitle

\begin{abstract}
Boosted trees is a dominant ML model, exhibiting high accuracy. 
However, boosted trees are hardly intelligible, and this is a problem whenever they are
used in safety-critical applications. Indeed, in such a context, rigorous  
explanations of the predictions made are expected. Recent work have shown how 
subset-minimal abductive explanations can be derived for boosted trees, using automated reasoning techniques.
However, the generation of such well-founded explanations is intractable in the general case.
To improve the scalability of their generation, we introduce the notion of tree-specific explanation for a
boosted tree. We show that tree-specific explanations are abductive explanations that can be
computed in polynomial time. We also explain how to derive a subset-minimal abductive explanation
from a tree-specific explanation. Experiments on various datasets show the computational benefits of
leveraging tree-specific explanations for deriving subset-minimal abductive explanations.
\end{abstract}

\section{Introduction}

The deployment of ML models in a large spectrum of applications has triggered the 
fast-growing development of eXplainable AI (XAI) (see for instance \cite{DBLP:conf/aiia/FrosstH17,GuidottiMRTGP19,Hooker19,Huysmans11,DBLP:conf/aaai/IgnatievNM19,Kim18,Lundberg17,DBLP:journals/ai/Miller19,Molnar19,DBLP:conf/sat/ShihDC19}).  
Models with high prediction performance are usually considered as poorly intelligible \cite{Molnar19,DBLP:journals/inffus/ArrietaRSBTBGGM20,DBLP:journals/natmi/LundbergECDPNKH20,DBLP:conf/kdd/CaruanaLRNJ20,DBLP:journals/corr/abs-2103-11251}.
Among them is the family of \emph{boosted trees} \cite{Friedman01}, which is among the state-of-the-art ML models when dealing with tabular 
data \cite{DBLP:journals/corr/abs-2110-01889}.

\paragraph{Related Work.} The design of efficient methods for interpreting ML models and explaining their decisions is acknowledged as an issue of 
the utmost importance when ML models are to be used in safety-critical applications \cite{MarquesSilva-Ignatiev22}.
Since most existing approaches to explaining ML models deliver model-agnostic explanations, they 
cannot be used in any high-risk context because the explanations that are generated are \emph{unsound}: one can find "counterexamples", i.e., instances that are 
covered by the same explanation but are nevertheless classified differently by the model \cite{DBLP:journals/corr/abs-1907-02509}.
Especially, \cite{DBLP:conf/ijcai/Ignatiev20} shows that the amount of counterexamples can be high when 
using some of the most popular approaches for computing model-agnostic explanations, namely LIME \cite{Lime16}, 
Anchors \cite{Anchor18}, and SHAP \cite{Lundberg17}.

In order to avoid the generation of unsound explanations, a number of alternative approaches, 
falling under the \emph{formal XAI} umbrella  \cite{MarquesSilva-Ignatiev22},
have shown how ML models of various types (including ``black'' boxes) can be associated with 
Boolean circuits (alias transparent or ``white'' boxes), exhibiting the same input-output behaviours  
(see among others  \cite{DBLP:conf/aaai/NarodytskaKRSW18,ShihChoiDarwiche18b,ShihChoiDarwiche19}). 
Thanks to such mappings, XAI queries about classifiers, including the generation of explanations, 
can be delegated to the corresponding circuits (see for instance \cite{DarwicheHirth20,DBLP:conf/nips/BarceloM0S20,DBLP:conf/icml/ParmentierV21}). 

Ensemble methods (bagging, boosting, stacking, etc.) have been considered in such a 
perspective. Thus, \cite{Choietal20,joao-ijcai21,Audemardetal22} show how to derive \emph{abductive explanations}
for random forests \cite{Breiman01}. An abductive explanation for an instance given a classifier is a subset of the characteristics of the instance that is enough to justify how the 
instance has been classified. In order to avoid the presence of useless characteristics in explanations, 
subset-minimal abductive explanations (alias sufficient reasons \cite{DarwicheHirth20}) are often targeted.
As to boosted trees, \cite{DBLP:journals/corr/abs-1907-02509} provides an SMT (satisfiability modulo theory) encoding scheme for
 boosted trees and shows how to use an SMT solver to compute sufficient reasons based on the encoding scheme. 
 The corresponding XAI tool is called {\tt XPlainer} (\url{https://github.com/alexeyignatiev/xplainer}).
\cite{Ignatievetal22} presents another encoding scheme, based on MaxSAT (maximum satisfiability), and indicates how to
 exploit a MaxSAT solver to compute sufficient reasons based on it. The associated tool is 
 called {\tt XReason} (\url{https://github.com/alexeyignatiev/xreason}). Deciding whether a given explanation is sound is intractable for boosted trees.
Accordingly, though {\tt XReason} typically exhibits better performances than {\tt XPlainer}, 
its scalability is still an issue. 

\paragraph{Contributions.} \emph{Showing how to enlarge the set of sufficient reasons that can be computed in practice 
for large datasets is the main goal of this paper.}
To reach this objective, we introduce the notion of \emph{tree-specific explanation} for a boosted tree. 
We show that, unlike sufficient reasons for boosted trees, tree-specific explanations are abductive explanations that can be
computed \emph{in polynomial time}. We also show that, while tree-specific explanations are not subset-minimal in the general case, they turn out to be  
close to sufficient reasons in practice. Furthermore, because sufficient reasons can be derived 
from tree-specific explanations, computing tree-specific explanations can be exploited as a preprocessing 
step in the derivation of sufficient reasons.
Experiments on various datasets show that leveraging tree-specific explanations for generating sufficient reasons 
is a valuable approach.

The proofs of the propositions presented in the paper are reported in a final appendix. A description of the datasets and the code used in our experiments are available from \url{www.cril.fr/expekctation/}.

\section{Preliminaries}\label{preliminaries}

For an integer $n$, let $[n] = \{1,\cdots,n\}$.
We consider a finite set $\{A_1, \ldots, A_n\}$ of \emph{attributes} (aka \emph{features}) where each attribute $A_i$ ($i \in [n]$) takes its value
in a domain $D_i$. Three types of attributes are taken into account: \emph{numerical} (the domain $D_i$ is a totally ordered set of numbers,
typically real numbers $\mathbb{R}$, or integers $\mathbb{Z}$), \emph{categorical} (the domain is a set of values that are not specifically ordered, 
e.g., $D_i = \{\mathit{b(lue)}, \mathit{w(hite)}, \mathit{r(ed)}\}$), or \emph{Boolean}
(the domain $D_i$ is $\mathbb{B} = \{0, 1\}$). An \emph{instance} $\vec x$ is a vector $(v_1, \ldots, v_n)$ where each $v_i$ ($i \in [n]$) 
is an element of $D_i$. $\vec x$ is also viewed as a term, i.e., a conjunctively-interpreted set of propositional atoms $t_{\vec x} = \{ (A_i = v_i) : i \in [n]\}$,
stating that each attribute $A_i$ takes the corresponding value $v_i$. 
Each pair $A_i = v_i$ is called a \emph{characteristic} of the instance. 
$\vec X$ denotes the set of all instances.

In the binary case, a \emph{classifier} $f$ is defined as a mapping from $\vec X$ to $\{1, 0\}$. When $f(\vec x) = 1$, 
$\vec x$ is said to be a positive instance, otherwise it is a negative instance. The set of all positive instances
forms a target concept, and the set of all negative instances is the complementary concept.
More generally, in the multi-class case, more than one concept (together with the complementary concept) is considered.
A \emph{classifier} $f$ is then defined as a mapping 
from $\vec X$ to $[m]$ with $m > 1$. Each integer from $[m]$ identifies a class and when 
$f(\vec x) = j$ with $j \in [m]$,  the instance $\vec x$ is said to be classified as an element of class $j$.

\paragraph{Trees and Forests.}
A \emph{regression tree} over $\{A_1, \ldots, A_n\}$ is a binary tree $T$, 
each of its internal nodes being labeled with a Boolean condition on an attribute from $\{A_1, \ldots, A_n\}$, 
and leaves are labeled by real numbers. The conditions are typically of the form $A_i > v_j$ with $v_j$ a number 
when $A_i$ is a numerical attribute, $A_i = v_j$ when $A_i$ is a categorical attribute, and $A_i$ (or equivalently $A_i = 1$)
when $A_i$ is a Boolean attribute.
The weight $w(T, \vec x) \in \mathbb{R}$ of $T$ for an input instance $\vec x \in \vec X$ is given by the label of the
leaf reached from the root as follows: at each node go to the left or right child depending 
on whether or not the condition labelling the node is satisfied by $\vec x$. A \emph{decision tree} over $\{A_1, \ldots, A_n\}$
is a regression tree over $\{A_1, \ldots, A_n\}$ where leaves are labeled in $\{0, 1\}$.

A \emph{forest} over $\{A_1, \ldots, A_n\}$ associated with a class $j \in [m]$ is an ensemble 
of trees $F^j = \{T_1^j,\cdots,T_{p_j}^j\}$, where each $T_k^j$ $(k \in [p_j])$ 
is a regression tree over $\{A_1, \ldots, A_n\}$, and such that the weight $w(F^j, \vec x) \in \mathbb{R}$ 
of $F^j$ for an input instance $\vec x \in \vec X$ is 
given by  $$w(F^j, \vec x) = \sum_{k=1}^{p_j} w(T_k^j, \vec x).$$ 
A \emph{random forest} over $\{A_1, \ldots, A_n\}$ is a forest over $\{A_1, \ldots, A_n\}$
that consists only of decision trees.

In a binary classification case, a \emph{boosted tree} $BT$ over $\{A_1, \ldots, A_n\}$ is a forest $F = \{T_1,\cdots,T_{p}\}$.
In a multi-class context, a \emph{boosted tree} $BT$ over $\{A_1, \ldots, A_n\}$ is a collection of $m$
forests $BT = \{F^1,\ldots, F^m\}$ over $\{A_1, \ldots, A_n\}$.
The size of a forest $F^j$ is given by $\size{F^j} = \sum_{k=1}^{p_j} \size{T_k^j}$,
where $\size{T_k^j}$ is the number of nodes occurring in $T_k^j$. 
The size of a boosted tree $BT$ is given by $\size{BT} = \sum_{j=1}^{m} \size{F^j}$.

In a binary classification case, an instance $\vec x$ is considered as a \emph{positive instance} when $w(F, \vec x) > 0$ and as a negative
instance otherwise. We note $BT(\vec x) = 1$ in the first case and $BT(\vec x) = 0$ in the second case.
In a multi-class context, an instance $\vec x$ is classified as an element of class
$j \in [m]$, noted $BT(\vec x) = j$, if and only if $w(F^j, \vec x) > w(F^i, \vec x)$ for every $i \in [m] \setminus \{j\}$.
If $w(F^j, \vec x) = w(F^i, \vec x)$ for every $i, j \in [m]$, then $BT(\vec x)$ is defined as a preset element of $[m]$ (e.g., a most
frequent class in the dataset used to learn $BT$).
Whatever the case (binary or multi-class), computing $BT(\vec x)$ can be achieved in polynomial time in $\size{BT}+n$.


   
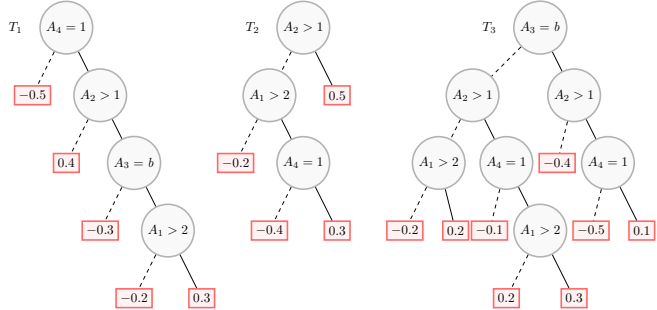
\begin{figure}[ht]
\centering
\scalebox{0.5}{
      \begin{tikzpicture}[scale=0.9, roundnode/.style={circle, draw=gray!60, fill=gray!5, very thick, minimum size=7mm},
      squarednode/.style={rectangle, draw=red!60, fill=red!5, very thick, minimum size=5mm}]
        \node[roundnode](root) at (2,7){$A_4 = 1$};
        \node at (0.5,7){$T_1$};
        \node[squarednode](n1) at (1,5){$-0.5$};
        \node[roundnode](n2) at (3,5){$A_2 > 1$};
        \node[squarednode](n21) at (2,3){$0.4$};
        \node[roundnode](n22) at (4,3){$A_3 = \mathit{b}$};  
        \node[squarednode](n221) at (3,1){$-0.3$};
        \node[roundnode](n222) at (5,1){$A_1 > 2$};   
        \node[squarednode](n2221) at (4,-1){$-0.2$};
        \node[squarednode](n2222) at (6,-1){$0.3$};
        \draw[dashed] (root) -- (n1);
        \draw(root) -- (n2);
        \draw[dashed] (n2) -- (n21);
        \draw(n2) -- (n22);    
        \draw[dashed] (n22) -- (n221);
        \draw(n22) -- (n222);  
        \draw[dashed] (n222) -- (n2221);
        \draw(n222) -- (n2222);
                
        \node[roundnode](root2) at (9,7){$A_2 > 1$};
        \node at (7.5,7){$T_2$};
        \node[roundnode](n2-1) at (8,5){$A_1 > 2$};  
        \node[squarednode](n2-2) at (10,5){$0.5$};
        \node[squarednode](n2-11) at (7,3){$-0.2$};
        \node[roundnode](n2-12) at (9,3){$A_4 = 1$};  
        \node[squarednode](n2-121) at (8,1){$-0.4$};
        \node[squarednode](n2-122) at (10,1){$0.3$};
        \draw[dashed] (root2) -- (n2-1);
        \draw(root2) -- (n2-2);
        \draw[dashed] (n2-1) -- (n2-11);
        \draw(n2-1) -- (n2-12);    
        \draw[dashed] (n2-12) -- (n2-121);
        \draw(n2-12) -- (n2-122);  
        
        \node[roundnode](root3) at (16,7){$A_3 = \mathit{b}$};
        \node at (14.5,7){$T_3$};
        \node[roundnode](n3-1) at (14,5){$A_2 > 1$};  
        \node[roundnode](n3-2) at (17,5){$A_2 > 1$};  
        \node[roundnode](n3-11) at (13,3){$A_1 > 2$};
        \node[roundnode](n3-12) at (15,3){$A_4 = 1$};   
        \node[squarednode](n3-121) at (14.5,1){$-0.1$};
        \node[roundnode](n3-122) at (16,1){$A_1 > 2$};  
        \node[squarednode](n3-1221) at (15,-1){$0.2$};
        \node[squarednode](n3-1222) at (17,-1){$0.3$};
        \node[squarednode](n3-21) at (16.5,3){$-0.4$};
        \node[roundnode](n3-22) at (18,3){$A_4 = 1$};     
        \node[squarednode](n3-111) at (12,1){$-0.2$};
        \node[squarednode](n3-112) at (13.5,1){$0.2$};
        \node[squarednode](n3-221) at (17.5,1){$-0.5$};
        \node[squarednode](n3-222) at (19,1){$0.1$};
        \draw[dashed] (root3) -- (n3-1);
        \draw(root3) -- (n3-2);
        \draw[dashed] (n3-1) -- (n3-11);
        \draw(n3-1) -- (n3-12);    
        \draw[dashed] (n3-2) -- (n3-21);
        \draw(n3-2) -- (n3-22); 
        \draw[dashed] (n3-11) -- (n3-111);
        \draw(n3-11) -- (n3-112);  
        \draw[dashed] (n3-22) -- (n3-221);
        \draw(n3-22) -- (n3-222);
        \draw[dashed] (n3-12) -- (n3-121);
        \draw(n3-12) -- (n3-122);
        \draw[dashed] (n3-122) -- (n3-1221);
        \draw(n3-122) -- (n3-1222);
      \end{tikzpicture}
 }
    \caption{A boosted tree $BT = \{F\}$ consisting of a single forest $F = \{T_1, T_2, T_3\}$.\label{fig:orchids}}
    \end{figure}

\begin{example}\label{running-ex}
As an example of binary classification, consider four attributes: $A_1$, $A_2$ are numerical, $A_3$ is categorical, and $A_4$ is Boolean.
The boosted tree $BT = \{F\}$ in Figure \ref{fig:orchids} is composed of a single forest $F$, which consists of three regression trees $T_1$, $T_2$, $T_3$. 

Consider $\vec x = (A_1 = 4, A_2 = 3, A_3 = b, A_4 = 1)$. We have $w(T_1, \vec x) = 0.3$, $w(T_2, \vec x) = 0.5$, and $w(T_3, \vec x) = 0.1$. Hence $w(F, \vec x) = 0.9$, and $\vec x$ is classified as a positive instance by $F$, thus it is classified as such by $BT$: $BT(\vec x) = 1$.

\end{example}

\paragraph{Abductive Explanations.}
Explaining the classification achieved by a classifier $f$ on an instance $\vec x$ consists in identifying a subset of the characteristics of $\vec x$
that is enough to get the class returned by $f$.
Formally, an \emph{abductive explanation} $t$ for an instance $\vec x \in \vec X$ given a classifier $f$ (that is binary or not) is a subset $t \in t_{\vec x}$
such that every instance $\vec x' \in \vec X$ such that $t \subseteq t_{\vec x'}$ is classified by $f$ in the same way as $\vec x$: $f(\vec x') = f(\vec x)$.\footnote{Unlike \cite{IgnatievNM19}, we do not require abductive explanations to be minimal w.r.t. set inclusion, in order to keep the concept 
distinct (and actually more general) than the one of sufficient reasons.} The size $|t|$ of an abductive explanation $t$ is the number of characteristics in it.
A \emph{sufficient reason} $t$ for $\vec x \in \vec X$ given $f$ is an abductive explanation for $\vec x$ given $f$ such that no proper subset $t'$ of $t$
is an abductive explanation for $\vec x$ given $f$. Stated otherwise, the sufficient reasons\footnote{Sufficient reasons are also known as prime-implicant explanations \cite{ShihCD18}.} for $\vec x$ given $f$ are the subset-minimal abductive explanations for $\vec x$ given $f$.

\begin{example} 
Considering again our running example, $t = \{(A_1 = 4),  (A_4 = 1)\}$ is a sufficient reason for $\vec x = (A_1 = 4, A_2 = 3, A_3 = b, A_4 = 1)$ given $BT = \{F\}$.
Indeed, all the instances $\vec x'$ extending $t$ can be gathered into four categories, obtained by considering the truth values of the Boolean conditions over the two remaining attributes ($A_2$ and $A_3$) as encountered in the trees of $BT$. In the four cases, we have $w(F, \vec x') > 0$ (see Table \ref{tab:sufficient}), showing that $BT(\vec x') = 1$. Since $BT(\vec x) = 1$, $t$ is an abductive explanation for $\vec x$ given $BT$. Since no proper subset of $t$ satisfies this property, $t$ actually is a
sufficient reason for $\vec x$ given $BT$.

\newcommand{\ra}[1]{\renewcommand{\arraystretch}{#1}}
\begin{table}[h]
\scalebox{0.9}{
\ra{0.9}
\begin{tabular}{cccc|ccc|c}
\midrule
$A_1= 4$ & $\mathbf{A_2 > 1}$ & $\mathbf{A_3 = b}$ & $A_4 = 1$ & $w(T_1, \vec x')$ & $w(T_2, \vec x')$ & $w(T_3, \vec x')$ & $w(F, \vec x')$\\
\midrule
$1$ & $\mathbf{0}$ & $\mathbf{0}$ & $1$ & $0.4$ & $0.3$ & $0.2$ & $0.9$\\
$1$ & $\mathbf{0}$ & $\mathbf{1}$ & $1$ & $0.4$ & $0.3$ & $-0.4$ & $0.3$\\
$1$ & $\mathbf{1}$ & $\mathbf{0}$ & $1$ & $-0.3$ & $0.5$ & $0.3$ & $0.5$\\
$1$ & $\mathbf{1}$ & $\mathbf{1}$ & $1$ & $0.3$ & $0.5$ & $0.1$ & $0.9$\\
\midrule
\end{tabular}
}
\caption{Weights of $BT$ for instances $\vec x'$ extending $t$.}\label{tab:sufficient}
\end{table}

\end{example}

Sufficient reasons are usually preferred to other abductive explanations since they are more simple: they do not contain any characteristics of
the instance at hand that are not useful to explain the prediction made by $f$.

%
%
%

\section{Computing Sufficient Reasons}\label{sec:sufficient}

In order to compute a sufficient reason for an input instance $\vec x$ given a classifier $f$, one can take advantage of a simple greedy algorithm (see Algorithm \ref{algo:greedy}). Starting with $t = t_{\vec x}$, this algorithm considers all the characteristics $c_i = (A_i = v_i)$ of $\vec x$ in a specific order and, at each step, tests whether 
$t$ deprived of $c_i$ is still an abductive explanation for $\vec x$ given $f$. If the test is positive, $c_i$ is removed from $t$, otherwise it 
is kept. Once all the characteristics $c_i$ of $\vec x$ have been considered, the resulting term $t$ is by construction a sufficient reason for $\vec x$ given $f$.

\begin{algorithm}[t]
\DontPrintSemicolon
$t \leftarrow t_{\vec x}$\\
\ForEach{$c_i \in t_{\vec x}$}
{
\lIf{$\texttt{implicant}(t \setminus \{c_i\}, \vec x, f)$}
{
$t \leftarrow t \setminus \{c_i\}$
}
}
\Return{$t$}
\caption{\texttt{SR}$(\vec x, f)$ \label{algo:greedy}}
\end{algorithm}

The computationally demanding step in this greedy algorithm is the call to function $\texttt{implicant}$ that tests whether 
$t$ deprived of $c_i$ is still an abductive explanation for $\vec x$ given $f$, i.e., any instance covered by $t \setminus \{c_i\}$
is classified in the same way as $\vec x$ by $f$. Though this test can be achieved in polynomial time 
for some families of classifiers $f$ (including decision trees) \cite{DBLP:journals/corr/abs-2010-11034,DBLP:conf/kr/HuangII021}, 
it is intractable in general. Especially, it is {\sf coNP}-hard when $f$ is a random forest \cite{Audemardetal22}.
Similarly, when $f$ is a boosted tree $BT$, we have:

\begin{proposition} \label{prop:complexityimplicanttestBT}
  Let $BT$ be a boosted tree over $\{A_1, \ldots, A_n\}$ and $\vec x \in \vec X$. Let $t$ be a subset of $t_{\vec x}$.
  Deciding whether $t$ is an abductive explanation for $\vec x$ given $BT$ is {\sf coNP}-complete.
  {\sf coNP}-hardness still holds in the restricted case every $A_i$ ($i \in [n]$) is Boolean and $BT$ consists of
  a single forest.
\end{proposition}

 In order to achieve the $\texttt{implicant}$ test when $f$ is a boosted tree $BT$, several approaches can be followed.
 \cite{DBLP:journals/corr/abs-1907-02509} took advantage of an SMT (SAT modulo theory) encoding of the
 boosted tree and then on an SMT solver to compute sufficient reasons. 
 More recently, \cite{Ignatievetal22} pointed out a more sophisticated encoding based on MaxSAT and 
 exploited a MaxSAT solver to compute sufficient reasons. Though this latter approach exhibited better performances
 in practice, its scalability is still an issue (the datasets considered in the experiments presented in \cite{Ignatievetal22} contain at most $60$ attributes).
\section{Computing Tree-Specific Explanations}

\paragraph{Worst / Best Instances.}
As explained before, when the classifier at hand is a regression tree, a forest, or (more generally) a boosted tree $BT$, the classification of an instance $\vec x \in \vec X$ depends on the weights of the tree(s) of the classifier for the instance. Because of this weight-based mechanism, the notion of abductive explanation $t$ for $\vec x$ can be characterized via the notion of \emph{worst /best instance} extending $t$. Let us start with the binary case:

\begin{definition}
Let $BT = \{F\}$ be a boosted tree over $\{A_1, \ldots, A_n\}$ and $\vec x \in \vec X$. Let $t$ be a subset of $t_{\vec x}$.
  \begin{itemize}
  \item A \emph{worst instance} extending $t$ given $F$ is an instance $\vec x' \in \vec X$ such that $t \subseteq t_{\vec x'}$ and
$w(F, \vec x') = \mathit{min}(\{w(F, \vec x'') : \vec x'' \in \vec X \mbox{ and } t \subseteq t_{\vec x''}\})$. 
  \item A \emph{best instance} extending $t$ given $F$ is an instance $\vec x' \in \vec X$ such that $t \subseteq t_{\vec x'}$ and
$w(F, \vec x') = \mathit{max}(\{w(F, \vec x'') : \vec x'' \in \vec X \mbox{ and } t \subseteq t_{\vec x''}\})$. 
\end{itemize}
 \end{definition}
 
 $W(t, F)$ (resp. $B(t, F)$) denotes the set of worst (resp. best) instances extending $t$ given $F$, and $w_\downarrow(t, F)$ (resp. $w_\uparrow(t, F)$) denotes the weight of any worst (resp. best) instance extending $t$ given $F$.
 
 On this ground, we have:
 
\begin{proposition} \label{prop:charact-worst}
In the binary case, let $BT = \{F\}$ be a boosted tree over $\{A_1, \ldots, A_n\}$ and $\vec x \in \vec X$. Let $t$ be a subset of $t_{\vec x}$.
\begin{itemize}
\item If $BT(\vec  x) = 1$, then $t$ is an abductive explanation for $\vec x$ given $BT$ if and only if any $\vec x' \in W(t, F)$ is such that $BT(\vec x') = 1$.
\item If $BT(\vec  x) = 0$, then $t$ is an abductive explanation for $\vec x$ given $BT$ if and only if any $\vec x' \in B(t, F)$ is such that $BT(\vec x') = 0$.
\end{itemize}
\end{proposition}

\begin{example} 
For our running example, $t = \{(A_1 = 4),  (A_4 = 1)\}$ is an abductive explanation for $\vec x = (A_1 = 4, A_2 = 3, A_3 = b, A_4 = 1)$ given $BT = \{F\}$ because 
any worst instance extending $t$, i.e., any $\vec x'$ satisfying $(A_1 = 4) \wedge  (A_2 \leq 1) \wedge  (A_3 = b) \wedge  (A_4 = 1)$ is such that
$w(F, \vec x') = 0.3$ (hence $w(F, \vec x') > 0$) (see Table \ref{tab:sufficient}). 
\end{example}

In the multi-class case, a similar notion of worst instance can be stated. \footnote{A notion of best instance could also be defined but it is useless for our purpose.}

\begin{definition}
  Let $BT = \{F^1, \ldots, F^m\}$ be a boosted tree over $\{A_1, \ldots, A_n\}$ and $\vec x \in \vec X$ such that $BT(\vec x) = i$. Let $t$ be a subset of $t_{\vec x}$.
 Given $BT$ and $\vec x$, a \emph{worst  instance} extending $t$ is an instance $\vec x' \in \vec X$ such that $t \subseteq t_{\vec x'}$ and $\vec x'$
  minimizes $w(F^i, \cdot) - \mathit{max}_{j \in [m] \setminus \{i\}} w(F^j, \cdot)$.
 \end{definition}
 
Then we have:
 
\begin{proposition} \label{prop:charact-worst-multiclass}
Let $BT = \{F^1, \ldots, F^m\}$ be a boosted tree over $\{A_1, \ldots, A_n\}$ and $\vec x \in \vec X$ such that $BT(\vec x) = i$. Let $t$ be a subset of $t_{\vec x}$.
$t$ is an abductive explanation for $\vec x$ given $BT$ if and only if for any worst instance $\vec x'$ extending $t$ given $BT$ and $\vec x$, we have 
$w(F^i, \vec x') - \mathit{max}_{j \in [m] \setminus \{i\}} w(F^j, \vec x') > 0$.
\end{proposition}


Propositions \ref{prop:complexityimplicanttestBT} and \ref{prop:charact-worst} (or \ref{prop:charact-worst-multiclass}) show together that identifying a worst (resp. best) instance $\vec x' \in \vec X$ extending a term $t \subseteq t_{\vec x}$ given a boosted tree $BT$ 
is intractable.\footnote{Intractability is still the case in the more specific case the classifier is a random forest.} Indeed,
if it were not the case, we could check in polynomial time
whether $t$ is an abductive explanation for $\vec x$ given $BT$ by testing whether $\vec x'$ is classified by $BT$ in the same way as $\vec x$.

\paragraph{Computing Worst/Best Instances for Trees.}
Interestingly, when the classifier consists of a regression tree $T$, identifying an element of $W(t, T)$ (resp. $B(t, T)$) is easy:
there exists a simple, linear-time, algorithm to compute $w_\downarrow(t, T)$  and $w_\uparrow(t, T)$, and
as a by-product, to derive a worst instance and a best instance extending $t$ given 
$T$. Basically, the algorithm consists of freezing in $T$ every arc corresponding to a condition
not satisfied by $t$, which can be done in time linear in the size of the input. A valid root-to-leaf 
path in the resulting tree is a root-to-leaf path of $T$ not containing any frozen arc. 
The weight $w_\downarrow(t, T)$ of $T$ for a worst (resp. best) instance extending $t$ simply is the minimal (resp. maximal) weight labelling a
leaf of a valid root-to-leaf path in the resulting tree, and it can be determined in time linear in the 
size of the input. Any $\vec x'$ satisfying the conditions associated with a valid root-to-leaf path
leading to a minimal (resp. maximal) weight leaf and satisfying $t \subseteq t_{\vec x'}$ is a worst (resp. best) 
instance extending $t$ given $T$.

\begin{example} 
Considering our running example again, let us identify worst instances extending $t = \{(A_1 = 4),  (A_4 = 1)\}$ for 
each of the trees $T_1$,  $T_2$, and $T_3$. On Figure \ref{fig:worst}, every frozen arc (and the corresponding subtree) is watermark displayed; 
the minimal weight leaves are bold, and the arcs of the corresponding root-to-leaf paths are bold. We have that:
\begin{itemize}
\item Every $\vec x' \in \vec X$ satisfying $(A_1 = 4) \wedge (A_2 > 1) \wedge (A_3 \neq b) \wedge (A_4 = 1)$ is an element of $W(t, T_1)$,
\item Every $\vec x' \in \vec X$ satisfying $(A_1 = 4) \wedge (A_2 \leq 1) \wedge (A_4 = 1)$ is an element of $W(t, T_2)$,
\item Every $\vec x' \in \vec X$ satisfying $(A_1 = 4) \wedge (A_2 \leq 1) \wedge (A_3 = b) \wedge (A_4 = 1)$ is an element of $W(t, T_3)$.
\end{itemize}

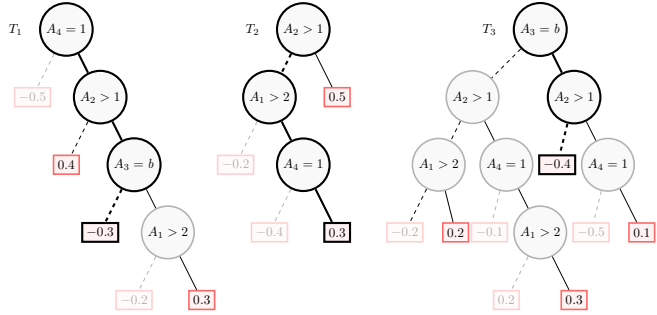
\begin{figure}[ht]
\centering
\scalebox{0.5}{
      \begin{tikzpicture}[scale=0.9, roundnode/.style={circle, draw=gray!60, fill=gray!5, very thick, minimum size=7mm},
      squarednode/.style={rectangle, draw=red!60, fill=red!5, very thick, minimum size=5mm}]
        \node[roundnode, draw=black, ultra thick](root) at (2,7){$A_4 = 1$};
        \node at (0.5,7){$T_1$};
        \node[squarednode, opacity=0.3](n1) at (1,5){$-0.5$};
        \node[roundnode, draw=black, ultra thick](n2) at (3,5){$A_2 > 1$};
        \node[squarednode](n21) at (2,3){$0.4$};
        \node[roundnode, draw=black, ultra thick](n22) at (4,3){$A_3 = \mathit{b}$};  
        \node[squarednode, draw=black, ultra thick](n221) at (3,1){$-0.3$};
        \node[roundnode](n222) at (5,1){$A_1 > 2$};   
        \node[squarednode,opacity=0.3](n2221) at (4,-1){$-0.2$};
        \node[squarednode](n2222) at (6,-1){$0.3$};
        \draw[dashed, opacity=0.3] (root) -- (n1);
        \draw[ultra thick](root) -- (n2);
        \draw[dashed] (n2) -- (n21);
        \draw[ultra thick](n2) -- (n22);    
        \draw[dashed, ultra thick] (n22) -- (n221);
        \draw(n22) -- (n222);  
        \draw[dashed,opacity=0.3] (n222) -- (n2221);
        \draw(n222) -- (n2222);
                
        \node[roundnode, draw=black, ultra thick](root2) at (9,7){$A_2 > 1$};
        \node at (7.5,7){$T_2$};
        \node[roundnode, draw=black, ultra thick](n2-1) at (8,5){$A_1 > 2$};  
        \node[squarednode](n2-2) at (10,5){$0.5$};
        \node[squarednode, opacity=0.3](n2-11) at (7,3){$-0.2$};
        \node[roundnode, draw=black, ultra thick](n2-12) at (9,3){$A_4 = 1$};  
        \node[squarednode, opacity=0.3](n2-121) at (8,1){$-0.4$};
        \node[squarednode, draw=black, ultra thick](n2-122) at (10,1){$0.3$};
        \draw[dashed, ultra thick] (root2) -- (n2-1);
        \draw(root2) -- (n2-2);
        \draw[dashed, opacity=0.3] (n2-1) -- (n2-11);
        \draw[ultra thick](n2-1) -- (n2-12);    
        \draw[dashed, opacity=0.3] (n2-12) -- (n2-121);
        \draw[ultra thick](n2-12) -- (n2-122);  
        
        \node[roundnode, draw=black, ultra thick](root3) at (16,7){$A_3 = \mathit{b}$};
        \node at (14.5,7){$T_3$};
        \node[roundnode](n3-1) at (14,5){$A_2 > 1$};  
        \node[roundnode, draw=black, ultra thick](n3-2) at (17,5){$A_2 > 1$};  
        \node[roundnode](n3-11) at (13,3){$A_1 > 2$};
        \node[roundnode](n3-12) at (15,3){$A_4 = 1$};   
        \node[squarednode, opacity=0.3](n3-121) at (14.5,1){$-0.1$};
        \node[roundnode](n3-122) at (16,1){$A_1 > 2$};  
        \node[squarednode, opacity=0.3](n3-1221) at (15,-1){$0.2$};
        \node[squarednode](n3-1222) at (17,-1){$0.3$};
        \node[squarednode, draw=black, ultra thick](n3-21) at (16.5,3){$-0.4$};
        \node[roundnode](n3-22) at (18,3){$A_4 = 1$};     
        \node[squarednode, opacity=0.3](n3-111) at (12,1){$-0.2$};
        \node[squarednode](n3-112) at (13.5,1){$0.2$};
        \node[squarednode, opacity=0.3](n3-221) at (17.5,1){$-0.5$};
        \node[squarednode](n3-222) at (19,1){$0.1$};
        \draw[dashed] (root3) -- (n3-1);
        \draw[ultra thick](root3) -- (n3-2);
        \draw[dashed] (n3-1) -- (n3-11);
        \draw(n3-1) -- (n3-12);    
        \draw[dashed, ultra thick] (n3-2) -- (n3-21);
        \draw(n3-2) -- (n3-22); 
        \draw[dashed] (n3-11) -- (n3-111);
        \draw(n3-11) -- (n3-112);  
        \draw[dashed, opacity=0.3] (n3-22) -- (n3-221);
        \draw(n3-22) -- (n3-222);
        \draw[dashed, opacity=0.3] (n3-12) -- (n3-121);
        \draw(n3-12) -- (n3-122);
        \draw[dashed, opacity=0.3] (n3-122) -- (n3-1221);
        \draw(n3-122) -- (n3-1222);
      \end{tikzpicture}
 }
    \caption{Worst instances and the corresponding weights for the regression trees used in $BT$.\label{fig:worst}}
    \end{figure}
\end{example}

\paragraph{Tree-Specific Explanations.}
We are now in position to define the notion of \emph{tree-specific explanation} $t$ for an instance $\vec x$ given a boosted tree $BT$.
We start with the binary classification case, i.e., when $BT$ consists of a single forest $F$:

\begin{definition}
  Let $F = \{T_1,\cdots,T_{p}\}$ be a forest over $\{A_1, \ldots, A_n\}$ and $\vec x \in \vec X$. 
  \begin{itemize}
  \item If $F(\vec x) = 1$, then $t$ is a \emph{tree-specific explanation} for $\vec x$ given $F$ if and only if $t$ is a subset of $t_{\vec x}$ such that 
  $\sum_{k=1}^p w_\downarrow(t, T_k) > 0$ and no proper subset of $t$ satisfies the latter condition.
  \item If $F(\vec x) = 0$, then $t$ is a \emph{tree-specific explanation} for $\vec x$ given $F$ if and only if $t$ is a subset of $t_{\vec x}$ such that 
  $\sum_{k=1}^p w_\uparrow(t, T_k) \leq 0$ and no proper subset of $t$ satisfies the latter condition.
   \end{itemize}
 \end{definition}

%
%
%

More generally, in the multi-class setting, tree-specific explanations can be defined as follows:

\begin{definition}
  Let $BT = \{F^1,\cdots, F^{m}\}$ be a boosted tree over $\{A_1, \ldots, A_n\}$ where each $F^j$ ($j \in [m]$) contains $p_j$ trees, 
  and $\vec x \in \vec X$ such that $BT(\vec x) = i$. 
  $t$ is a \emph{tree-specific explanation} for $\vec x$ given $BT$ if and only if $t$ is a subset of $t_{\vec x}$ such that for every $j \in [m] \setminus \{i\}$, 
  we have $\sum_{k=1}^{p_i} w_\downarrow(t, T_k^i) > \sum_{k=1}^{p_j} w_\uparrow(t, T_k^j)$, and no proper subset of $t$ satisfies the latter condition.
 \end{definition}


A first key property that makes tree-specific explanations valuable is that they are abductive explanations:

\begin{proposition} \label{prop:specific-abductive}
Let $BT$ be a boosted tree over $\{A_1, \ldots, A_n\}$ and $\vec x \in \vec X$. 
If $t$ is a tree-specific explanation for $\vec x$ given $BT$, then 
$t$ is an abductive explanation for $\vec x$ given $BT$.
\end{proposition}

Especially, each time the test $\forall j \in [m] \setminus \{i\}, \sum_{k=1}^{p_i} w_\downarrow(t, T_k^i) > \sum_{k=1}^{p_j} w_\uparrow(t, T_k^j)$
succeeds, it is ensured that $t$ is an abductive explanation for $\vec x$ given $BT$. However, the condition is only sufficient: when the test fails, it can be the case that
$t$ is an abductive explanation for $\vec x$ given $BT$ nevertheless. Testing the condition $\forall j \in [m] \setminus \{i\}, \sum_{k=1}^{p_i} w_\downarrow(t, T_k^i, \vec x) > \sum_{k=1}^{p_j} w_\uparrow(t, T_k^j, \vec x)$ thus amounts to making an \emph{incomplete implicant test}.

It is easy to check that tree-specific explanations coincide with sufficient reasons for regression trees. Unsurprisingly, given the
complexity shift pointed out in Proposition \ref{prop:complexityimplicanttestBT},
this equivalence does not hold for forests or boosted trees. 
Thus, in the general case, a tree-specific explanation $t$ for $\vec x$ given $BT$ is not a 
sufficient reason for $\vec x$ given $BT$: $t$ may contain characteristics of $\vec x$ that could be removed
without questioning the classification achieved by $BT$. 

\begin{example} 
Considering our running example again, the sufficient reason $t = \{(A_1 = 4),  (A_4 = 1)\}$ for $\vec x = (A_1 = 4, A_2 = 3, A_3 = b, A_4 = 1)$ given $BT = \{F\}$
is not a tree-specific explanation for $\vec x$ given $BT$. Indeed, we have 
$w_\downarrow(t, T_1) = -0.3$, $w_\downarrow(t, T_2) = 0.3$, and $w_\downarrow(t, T_3) = -0.4$, hence 
$w_\downarrow(t', T_1) + w_\downarrow(t', T_2) + w_\downarrow(t', T_3) = -0.4 < 0$ while $w(F, \vec x) = 0.9 > 0$.
Contrastingly, $t' = \{(A_2 = 3),  (A_4 = 1)\}$  is a tree-specific explanation for $\vec x$ given $BT$.
We have $w_\downarrow(t', T_1) = -0.3$, $w_\downarrow(t', T_2) = 0.5$, and $w_\downarrow(t', T_3) = 0.1$, hence 
$w_\downarrow(t', T_1) + w_\downarrow(t', T_2) + w_\downarrow(t', T_3) = 0.3 > 0$.
It can be verified that $t'$ also is a  sufficient reason for $\vec x$ given $BT$.
\end{example} 

Though subset-minimality is required in both cases, the discrepancy between tree-specific explanations and sufficient reasons
can be easily explained by the fact that tree-specific explanations consider the trees \emph{separately}: it can be easily the case
that two distinct trees $T_k^ j$ and $T_l^j$ belonging to the same forest $F^j$ do not share any worst instance extending a given term $t$.
In symbols, we may have $W(t, T_k^j) \cap W(t, T_l^j) = \emptyset$. 

\begin{example} 
For our running example, no worst instance extending $t = \{(A_1 = 4),  (A_4 = 1)\}$ given $T_1$ is also a 
worst instance extending $t = \{(A_1 = 4),  (A_4 = 1)\}$ given $T_2$ or given $T_3$. Indeed, 
every worst instance extending $t = \{(A_1 = 4),  (A_4 = 1)\}$ given $T_1$ must satisfy $A_2 > 1$, while  
every worst instance extending $t = \{(A_1 = 4),  (A_4 = 1)\}$ given $T_2$ or $T_3$ must satisfy
the complementary condition $A_2 \leq 1$.
\end{example} 

In the worst case, the number of useless characteristics in a tree-specific explanation can be equal to the number $n$ of attributes:

\begin{proposition}\label{prop:discrepancy}
Let $BT$ be a boosted tree over $\{A_1, \ldots, A_n\}$ and $\vec x \in \vec X$. It can be the case that 
the unique tree-specific explanation for $\vec x$ given $BT$ consists of $t_{\vec x}$ itself, while $\emptyset$
is the unique sufficient reason for $\vec x$ given $BT$. This holds even in the restricted case $BT$ consists of a single forest
and every attribute is Boolean.
\end{proposition}

A second key property that makes tree-specific explanations valuable is that they can be computed efficiently. 
Thus, the greedy algorithm \texttt{TS} given by Algorithm \ref{algo:tree-specific} 
can be used to derive in $\mathcal{O}(n \size{BT})$ a tree-specific explanation for $\vec x$ given $BT$ in the multi-class case. 

\begin{algorithm}[t]
\DontPrintSemicolon
\SetKw{Break}{break}
$t \leftarrow t_{\vec x}$\\
$j \leftarrow BT(\vec x)$\\
\ForEach{$c_i \in t_{\vec x}$}
{
\If{$\nexists k \in [m] \setminus \{j\} \text{ s.t. } \sum_{l=1}^{p_j} w_\downarrow(t \setminus \{c_i\}, T_l^j) \leq \sum_{l=1}^{p_k} w_\uparrow(t \setminus \{c_i\}, T_l^k)$}{
$t \leftarrow t \setminus \{c_i\}$
}
}
\Return{$t$}
\caption{\texttt{TS}$(\vec x, BT)$ \label{algo:tree-specific}}
\end{algorithm}

\begin{proposition}\label{prop:correctness}
Let $BT$ be a boosted tree over $\{A_1, \ldots, A_n\}$ and $\vec x \in \vec X$. 
\texttt{TS}$(\vec x, BT)$ returns a tree-specific explanation for $\vec x$ given $BT$.
\end{proposition}

Clearly enough, an algorithm closely similar to \texttt{TS} can be designed to handle the binary classification case
(in that case, at each iteration, one just needs to test the sign of $\sum_{k=1}^p w_\downarrow(t, T_k)$ when the instance is positive, and
the sign of $\sum_{k=1}^p w_\uparrow(t, T_k)$ when the instance is negative).

Interestingly, when dealing with boosted trees, the greedy algorithm \texttt{SR} (Algorithm \ref{algo:greedy}) 
for deriving sufficient reasons can be exploited to remove useless characteristics in
tree-specific explanations, i.e., to generate sufficient reasons from tree-specific explanations. Viewed from a different angle,
the computation of a tree-specific explanation for an instance $\vec x$ given a boosted tree $BT$ can be exploited as a \emph{preprocessing} step
in \texttt{SR}. This combination is given by the pipeline $\texttt{SR}(\texttt{TS}(t_{\vec x}, BT), BT)$.



The rationale for this preprocessing step is the fact that \texttt{TS} is a polynomial-time algorithm, while \texttt{implicant} is not.
As the experiments reported in Section \ref{experiments} will show it, \texttt{TS} may remove in a very efficient way 
many useless characteristics of $\vec x$, thus avoiding many calls to the computationally expensive function \texttt{implicant}.

\section{Experiments}\label{experiments}

\paragraph{Empirical Protocol.}
The empirical protocol was as follows.
We have considered 50 datasets, which are standard benchmarks ({adult, farm-ads}...) coming from the well-known repositories Kaggle (\url{www.kaggle.com}), OpenML (\url{www.openml.org}), and UCI (\url{archive.ics.uci.edu/ml/}).
For these datasets, the number of classes varies from $2$ to $9$ classes, the number of attributes (features) from $10$ to $100 001$, and the number of instances from $345$ to $48 842$.
Categorical features have been treated as numbers. 
As to numerical features, no data preprocessing has taken place: these features have been binarized on-the-fly by the learning algorithm that has been used, namely
XGBoost \cite{Chen16} that learns gradient boosted trees. All parameters have been set to their default values (especially, $100$ trees per class have been 
considered and the maximum depth of each tree was set to $6$). 

For every dataset, a $10$-fold cross validation process has been achieved.
Ten boosted trees have been learned per dataset. The mean accuracy per dataset varies from $53.23\%$ up to $100\%$ (in average, it is equal to $88.5\%$).
Ten instances have been picked up uniformly at random in the test set associated with the training set used to
learn each boosted tree. This led to $100$ instances per dataset, giving a total of $5000$ instances for which
explanations about the way they were classified have been sought for. To get such explanations,
we ran implementations of the algorithms presented in the previous sections: {\tt SR} implemented as 
{\tt XReason} (with its default parameters), our own implementation of {\tt TS}, and an implementation
{\tt TS}+{\tt XReason} of the pipeline of the two. In order to implement this pipeline, we had to
modify {\tt XReason} in such a way that it can use as input any abductive explanation for the instance at hand, and not only
 the instance itself.  By default, {\tt XReason} starts by removing some useless characteristics of the input instance
using a core-guided mechanism. Though beneficial when {\tt XReason} is used alone, this step turns out to be counter-productive
when  {\tt XReason} is combined with {\tt TS}. Indeed, in our experiments, the number of instances (out of $5000$) 
"solved" by {\tt TS}+{\tt XReason} with this treatment switched on 
 is $4016$, while it is equal to $4097$ when the treatment was switched off. Hence, in our experiments, the treatment has been switched off when 
  {\tt TS} was run upstream to {\tt XReason}, and switched on when {\tt XReason} was used alone.
%
We have also modified {\tt XReason} to make it provide the abductive explanation that is available when the time limit is reached (in this case, 
the returned explanation is not guaranteed to be subset-minimal).
In our experiments, for each instance $\vec x$, {\tt TS} has been called
1000 times: at each run, an elimination ordering of the characteristics of $\vec x$ (as considered at line 3. of Algorithm \ref{algo:tree-specific}) has been picked up 
uniformly at random, and a shortest tree-specific explanation among those generated for the 1000 runs was finally returned. 

All the experiments have been conducted on a computer equipped with Intel(R) XEON E5-2637 CPU @ 3.5 GHz and 128 Gib of memory.
For each algorithm, a timeout (TO) of 100 seconds per instance has been considered. 

\paragraph{Results.}

\begin{figure}[h]
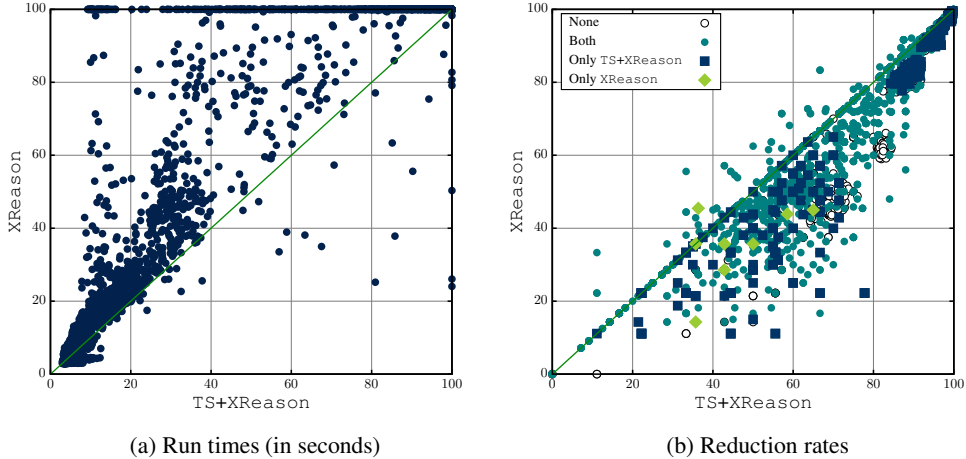

     \centering
     
     \hspace*{0cm}\begin{subfigure}[b]{0.4\textwidth}
         {\graphicspath{{figures/SCATTERTIMES/}}
            \scalebox{.52}{\input{figures/SCATTERTIMES/cactus.tex}}
         }
         \vspace*{-0.3cm}\hspace*{1cm}\caption{Run times (in seconds)}
     \end{subfigure}
     \hspace*{1cm}\begin{subfigure}[b]{0.4\textwidth}
         {\graphicspath{{figures/SCATTERREDUC/}}
            \scalebox{.52}{\input{figures/SCATTERREDUC/cactus.tex}}
         }
         \vspace*{-0.3cm}\hspace*{1cm}\caption{Reduction rates}
     \end{subfigure}
     
     \caption{Comparing {\tt TS}+{\tt XReason} to {\tt XReason}.}
     \label{fig:scatter}
\end{figure}


A synthesis of the results we obtained is provided on Figure~\ref{fig:scatter}. On those two scatter plots, each dot 
corresponds to an instance among the $5000$ instances tested. 

Figure~\ref{fig:scatter} (a) is about computation times.
The $x$-coordinate (resp. $y$-coordinate) of a dot is the time (in seconds)  required by {\tt TS}+{\tt XReason} 
(resp. {\tt XReason}) to compute an abductive explanation for the associated instance. By construction, this abductive explanation is 
a sufficient reason for the instance when the computation stops before the time limit. In light of this scatter plot, two observations can be made.
On the one hand, the run times of {\tt TS}+{\tt XReason} are significantly smaller than those of {\tt XReason}.
On the other hand, the time limit has been reached much more often by {\tt XReason} than by {\tt TS}+{\tt XReason}.

%

Figure~\ref{fig:scatter} (b) is about the size of the abductive explanations that are generated, and more precisely, about reduction rates. 
Indeed, size is one of the criteria to be considered when evaluating the intelligibility\footnote{In general, the intelligibility of an explanation 
does not reduce to its size and an accurate evaluation of it cannot be achieved in a context-independent way \cite{DoshiVelezK17,DBLP:journals/corr/abs-1802-00682}, since intelligibility typically depends on the explainee (i.e., the person who asked for an explanation) \cite{Miller19}.} of an explanation: everything else being equal, shortest explanations are easier to understand than longer explanations. The reduction rate achieved by an abductive explanation $t$ for an instance $\vec x = (v_1, \ldots, v_n)$
is given by $1 - \frac{|t|}{n}$.
For each dot corresponding to an instance $\vec x = (v_1, \ldots, v_n)$, the $x$-coordinate of the dot is the reduction rate achieved by the abductive explanation generated by {\tt TS}+{\tt XReason} for $\vec x$, while the $y$-coordinate is the reduction rate achieved by the abductive explanation generated by {\tt XReason} for $\vec x$.
We can observe on this figure that {\tt TS}+{\tt XReason} produces in general much smaller abductive explanations than those generated by {\tt XReason}.

On Figure~\ref{fig:scatter} (b), we used different dot representations for instances, depending on the fact that a sufficient reason for the instance at hand 
has been computed (or not) within the time limit by any of the two programs, or by both of them.
One can observe that the number of sufficient reasons that have been (provably) derived by {\tt TS}+{\tt XReason} in at most $100$ seconds 
is significantly higher than the number of sufficient reasons that have been (provably) derived by {\tt XReason}. More in detail, 
out of $5000$ abductive explanations, $3476$ sufficient reasons have been obtained in due time by the two programs, while 
$621$ have been obtained in due time by {\tt TS}+{\tt XReason} alone, $8$ have been obtained in due time by {\tt XReason} alone, and
for $895$ abductive explanations that have been generated, there are no subset-minimality guarantees for any of the programs used.
Thus, a significant amount of $621$ sufficient reasons have been gained by taking advantage of {\tt TS} as a preprocessing to {\tt XReason}.


\setlength{\tabcolsep}{2.5pt}
\newcommand{\ra}[1]{\renewcommand{\arraystretch}{#1}}
\begin{table}[h]
  \centering
    \centering
   	\centering\ra{0.9}
        \renewcommand{\arraystretch}{0.9} 
	\scalebox{0.725}{    
    	\begin{tabular}[b]{l|rrrr|rrrr|rr|rr|r}	
        \toprule
      	\textbf{}\;\; & \multicolumn{4}{c|}{} & \multicolumn{6}{c|}{\textbf{Run time}} & \multicolumn{3}{c}{\textbf{Reduction rate}}\\
        \textbf{Dataset}\;\; & \textbf{\#Cls.} & \textbf{\#Feat.} & \textbf{\#Inst.} & \textbf{Acc.} & \multicolumn{4}{c|}{\textbf{{\tt TS}+{\tt XReason}}} & \multicolumn{2}{c|}{\textbf{{\tt XReason}}} & \multicolumn{2}{c|}{\textbf{{\tt TS}+{\tt XReason}}} & \multirow{2}{*}{\textbf{{\tt XReason}}}\\
        
      	\textbf{}\;\; & \multicolumn{4}{c|}{} & \textbf{{\tt TS}} & \textbf{{\tt XReason}} & \textbf{{\tt TS}+{\tt XReason}} & \multicolumn{1}{c|}{\textbf{{\tt \#TO}}} & \textbf{{\tt XReason}} & \textbf{{\tt \#TO}} & \textbf{{\tt TS}} & \textbf{{\tt TS}+{\tt XReason}} & \\
        \midrule
        gina\_agnostic & 2 & 971 & 3468 & 95.13	& 1.80 & 95.92 & 97.72 & 66 & 100 & 100 & 87.51 & 88.56 & 81.66 \\
        malware & 2 & 1085 & 6248 & 99.46 & 0.22 & 5.79 & 6.01 & 0 & 8.34 & 0 & 98.74 &	98.85 & 98.64 \\
        ad\_data & 2 & 1559 & 3279 & 97.78 & 0.32 & 13.88 & 14.20 & 0 & 71.41 & 47 & 98.95 & 99.18 & 98.26\\
        christine & 2 & 1637 & 5418 & 73.81 & 6.13 & 94.42 & 100 & 100 & 100 & 100 & 82.34 & 82.42 & 62.44\\
        cnae & 9 & 857 & 1079 & 91.48 & 5.65 & 49.57 & 55.23 & 3 & 89 & 50 & 94.17 & 94.90 & 92.61\\
        gisette & 2 & 5001 & 7000 & 97.83 & 2.95 & 90.59 & 93.54 & 23 & 100 & 100 & 97.05 & 97.23 & 95.46\\
        arcene & 2 & 10001 & 200 & 80.50 & 0.20 & 5.70 & 5.89 & 0 & 4.49 & 0 & 99.62 & 99.63 & 99.59\\
        dexter & 2 & 20001 & 600 & 91.83 & 0.30 & 6.75 & 7.05 & 0 & 11.25 & 0 & 99.88 & 99.90 & 99.85\\
        allBooks & 8 & 8267 & 590 & 87.12 & 7.62 & 51.01 & 58.62 & 3 & 100 & 100 & 99.12 & 99.23 & 98.54\\
        farm-ads & 2 & 54877 & 4143 & 90.27 & 2.86 & 97.91 & 99.99 & 93 & 100 & 100 & 99.62 & 99.64 & 99.55\\
        dorothea & 2 & 100001 & 1150 & 93.91 & 0.68 & 11.50 & 12.17 & 0 & 17.28 & 0 & 99.91 & 99.92 & 99.90\\
        \bottomrule
        \end{tabular}
        } 
  \caption{\label{tab:dataset} Performances of {\tt TS}+{\tt XReason} and {\tt XReason} in terms of run times and reduction rates for $10$ datasets.}
\end{table}

To complete the results furnished by the scatter plots, Table \ref{tab:dataset} presents some details concerning $10$ datasets among the $50$ datasets used.
The columns give, from left to right, the name of the dataset, the number of classes, features and instances in it, the mean accuracy, and  results (over $100$ instances) about {\tt TS}+{\tt XReason} and {\tt XReason} in terms of mean run times (in seconds)\footnote{Whenever a TO occurs, we considered a computation time equal to 100s.}, number of TOs, and then in terms of mean reduction rates. 
For {\tt TS}+{\tt XReason}, we report the mean run times and mean reduction rates achieved by each component of it (i.e.,  by {\tt TS} as a preprocessing step, and then by {\tt XReason} run on the abductive explanation generated by {\tt TS}).
In light of those results, {\tt TS} appears as valuable: {\tt TS} is computationally efficient (the cumulated run times over 1000 runs are bounded by a few seconds), and  {\tt TS} achieves important reduction rates (close to those achieved by {\tt TS}+{\tt XReason}). Notably, the reduction rates achieved by {\tt TS} are often higher than those achieved by {\tt XReason}. This explains the good performance of the pipeline {\tt TS}+{\tt XReason}. 

%


%



\section{Conclusion}\label{conclusion}

We have introduced a new notion of abductive explanation for boosted trees, called tree-specific explanations. 
In the worst case, tree-specific explanations can be arbitrarily larger than sufficient reasons, thus they may contain many useless
characteristics. However, we have shown that, unlike sufficient reasons, their generation is tractable.
We have presented a polynomial-time algorithm {\tt TS} for computing tree-specific explanations, and we have proved its correctness.
Because a sufficient reason can be extracted from a tree-specific explanation, {\tt TS} can be used
as a preprocessing step for greedy algorithms (like {\tt XReason}) that derive sufficient reasons from boosted trees. 
Empirically, we have shown that the combination {\tt TS}+{\tt XReason} significantly improves the state-of-the-art.
Finally, in practice, the abductive explanations computed by {\tt TS} are often close to sufficient reasons. This shows that
{\tt TS} can also be useful alone, as a generator of abductive explanations.

Many research perspectives can be envisioned. Instead of considering the characteristics
of the input instance randomly (line 3 of {\tt TS}), it would make sense to design heuristics for making 
more informed choices, incorporating domain knowledge about the characteristics. 
It would also be useful to investigate whether the notions of worst/best instances for a
forest could be exploited for improving the generation of counterfactual explanation
for boosted trees.

\section*{Acknowledgments} 
This work has benefited from the support of 
the AI Chair EXPE\textcolor{orange}{KC}TATION (ANR-19-CHIA-0005-01) of the French National Research Agency.
It was also partially supported by TAILOR, a project funded by EU Horizon 2020 research
and innovation programme under GA No 952215.

\bibliographystyle{plain}

\begin{thebibliography}{10}

\bibitem{DBLP:journals/inffus/ArrietaRSBTBGGM20}
A.~Barredo Arrieta, N.~D{\'{\i}}az R., J.~Del Ser, A.~Bennetot, S.~Tabik,
  A.~Barbado, S.~Garc{\'{\i}}a, S.~Gil{-}Lopez, D.~Molina, R.~Benjamins,
  R.~Chatila, and F.~Herrera.
\newblock Explainable artificial intelligence {(XAI):} concepts, taxonomies,
  opportunities and challenges toward responsible {AI}.
\newblock {\em Inf. Fusion}, 58:82--115, 2020.

\bibitem{Audemardetal22}
G.~Audemard, S.~Bellart, L.~Bounia, F.~Koriche, J.-M. Lagniez, and P.~Marquis.
\newblock Trading complexity for sparsity in random forest explanations.
\newblock In {\em Proc. of AAAI'22}, 2022.

\bibitem{DBLP:conf/nips/BarceloM0S20}
P.~Barcel{\'{o}}, M.~Monet, J.~P{\'{e}}rez, and B.~Subercaseaux.
\newblock Model interpretability through the lens of computational complexity.
\newblock In {\em Advances in Neural Information Processing Systems 33: Annual
  Conference on Neural Information Processing Systems 2020, NeurIPS 2020,
  December 6-12, 2020, virtual}, 2020.

\bibitem{DBLP:journals/corr/abs-2110-01889}
V.~Borisov, T.~Leemann, K.~Se{\ss}ler, J.~Haug, M.~Pawelczyk, and G.~Kasneci.
\newblock Deep neural networks and tabular data: {A} survey.
\newblock {\em CoRR}, abs/2110.01889, 2021.

\bibitem{Breiman01}
L.~Breiman.
\newblock Random forests.
\newblock {\em Machine Learning}, 45(1):5--32, 2001.

\bibitem{DBLP:conf/kdd/CaruanaLRNJ20}
R.~Caruana, S.~M. Lundberg, M.~T{\'{u}}lio Ribeiro, H.~Nori, and S.~Jenkins.
\newblock Intelligible and explainable machine learning: Best practices and
  practical challenges.
\newblock In Rajesh Gupta, Yan Liu, Jiliang Tang, and B.~Aditya Prakash,
  editors, {\em {KDD} '20: The 26th {ACM} {SIGKDD} Conference on Knowledge
  Discovery and Data Mining, Virtual Event, CA, USA, August 23-27, 2020}, pages
  3511--3512. {ACM}, 2020.

\bibitem{Chen16}
T.~Chen and C.~Guestrin.
\newblock {XGBoost}: A scalable tree boosting system.
\newblock In {\em Proc. of KDD'16}, page 785–794, 2016.

\bibitem{Choietal20}
A.~Choi, A.~Shih, A.~Goyanka, and A.~Darwiche.
\newblock On symbolically encoding the behavior of random forests.
\newblock In {\em Proc. of FoMLAS'20, 3rd Workshop on Formal Methods for
  ML-Enabled Autonomous Systems, Workshop at CAV'20}, 2020.

\bibitem{DarwicheHirth20}
A.~Darwiche and A.~Hirth.
\newblock On the reasons behind decisions.
\newblock In {\em Proc. of ECAI'20}, pages 712--720, 2020.

\bibitem{DoshiVelezK17}
F.~Doshi{-}Velez and B.~Kim.
\newblock A roadmap for a rigorous science of interpretability.
\newblock {\em CoRR}, abs/1702.08608, 2017.

\bibitem{Friedman01}
J.~H. Friedman.
\newblock Greedy function approximation.
\newblock {\em The Annals of Statistics: A Gradient Boosted Machine},
  29(5):1189--1232, 2001.

\bibitem{DBLP:conf/aiia/FrosstH17}
N.~Frosst and G.~E. Hinton.
\newblock Distilling a neural network into a soft decision tree.
\newblock In {\em Proc. of the First International Workshop on
  Comprehensibility and Explanation in {AI} and {ML}}, volume 2071 of {\em
  {CEUR} Workshop Proceedings}. CEUR-WS.org, 2017.

\bibitem{GuidottiMRTGP19}
R.~Guidotti, A.~Monreale, S.~Ruggieri, F.~Turini, F.~Giannotti, and
  D.~Pedreschi.
\newblock A survey of methods for explaining black box models.
\newblock {\em {ACM} Computing Surveys}, 51(5):93:1--93:42, 2019.

\bibitem{Hooker19}
S.~Hooker, D.~Erhan, P-J. Kindermans, and B.~Kim.
\newblock A benchmark for interpretability methods in deep neural networks.
\newblock In {\em Proc. of NeurIPS'19}, pages 9737--9748, 2019.

\bibitem{DBLP:conf/kr/HuangII021}
X.~Huang, Y.~Izza, A.~Ignatiev, and J.~Marques{-}Silva.
\newblock On efficiently explaining graph-based classifiers.
\newblock In {\em Proceedings of the 18th International Conference on
  Principles of Knowledge Representation and Reasoning, {KR} 2021, Online
  event, November 3-12, 2021}, pages 356--367, 2021.

\bibitem{Huysmans11}
J.~Huysmans, K.~Dejaeger, C.~Mues, J.~Vanthienen, and B.~Baesens.
\newblock An empirical evaluation of the comprehensibility of decision table,
  tree and rule based predictive models.
\newblock {\em Decis. Support Syst.}, 51(1):141–154, 2011.

\bibitem{DBLP:conf/ijcai/Ignatiev20}
A.~Ignatiev.
\newblock Towards trustable explainable {AI}.
\newblock In {\em Proceedings of the Twenty-Ninth International Joint
  Conference on Artificial Intelligence, {IJCAI} 2020}, pages 5154--5158.
  ijcai.org, 2020.

\bibitem{Ignatievetal22}
A.~Ignatiev, Y.~Izza, P.J. Stuckey, and J.~Marques-Silva.
\newblock Using {MaxSAT} for efficient explanations of tree ensembles.
\newblock In {\em Proc. of AAAI'22}, 2022.

\bibitem{DBLP:conf/aaai/IgnatievNM19}
A.~Ignatiev, N.~Narodytska, and J.~Marques{-}Silva.
\newblock Abduction-based explanations for machine learning models.
\newblock In {\em Proc. of AAAI'19}, pages 1511--1519, 2019.

\bibitem{IgnatievNM19}
A.~Ignatiev, N.~Narodytska, and J.~Marques{-}Silva.
\newblock Abduction-based explanations for machine learning models.
\newblock In {\em Proceedings of the 23rd {AAAI} Conference on Artificial
  Intelligence (AAAI'19)}, pages 1511--1519, 2019.

\bibitem{DBLP:journals/corr/abs-1907-02509}
A.~Ignatiev, N.~Narodytska, and J.~Marques{-}Silva.
\newblock On validating, repairing and refining heuristic {ML} explanations.
\newblock {\em CoRR}, abs/1907.02509, 2019.

\bibitem{DBLP:journals/corr/abs-2010-11034}
Y.~Izza, A.~Ignatiev, and J.~Marques{-}Silva.
\newblock On explaining decision trees.
\newblock {\em CoRR}, abs/2010.11034, 2020.

\bibitem{joao-ijcai21}
Y.~Izza and J.~Marques-Silva.
\newblock On explaining random forests with {SAT}.
\newblock In {\em Proceedings of the 30th International Joint Conference on
  Artificial Intelligence (IJCAI'21)}, 2021.

\bibitem{Kim18}
B.~Kim, M.~Wattenberg, J.~Gilmer, C.~Cai, J.~Wexler, F.~Viegas, and R.~Sayres.
\newblock Interpretability beyond feature attribution: Quantitative testing
  with concept activation vectors ({TCAV}).
\newblock In {\em Proc. of ICML'18}, pages 2668--2677, 2018.

\bibitem{Lundberg17}
S.~Lundberg and S-I. Lee.
\newblock A unified approach to interpreting model predictions.
\newblock In {\em Proc. of NIPS'17}, pages 4765--4774, 2017.

\bibitem{DBLP:journals/natmi/LundbergECDPNKH20}
S.~M. Lundberg, G.~G. Erion, H.~Chen, A.~J. DeGrave, J.~M. Prutkin, B.~Nair,
  R.~Katz, J.~Himmelfarb, N.~Bansal, and S.I. Lee.
\newblock From local explanations to global understanding with explainable {AI}
  for trees.
\newblock {\em Nat. Mach. Intell.}, 2(1):56--67, 2020.

\bibitem{MarquesSilva-Ignatiev22}
J.~Marques-Silva and A.~Ignatiev.
\newblock Delivering trustworthy {AI} through formal {XAI}.
\newblock In {\em Proc. of AAAI'22}, 2022.

\bibitem{DBLP:journals/ai/Miller19}
T.~Miller.
\newblock Explanation in artificial intelligence: Insights from the social
  sciences.
\newblock {\em Artificial Intelligence}, 267:1--38, 2019.

\bibitem{Miller19}
T.~Miller.
\newblock Explanation in artificial intelligence: Insights from the social
  sciences.
\newblock {\em Artificial Intelligence}, 267:1--38, 2019.

\bibitem{Molnar19}
Ch. Molnar.
\newblock {\em Interpretable Machine Learning - A Guide for Making Black Box
  Models Explainable}.
\newblock Leanpub, 2019.

\bibitem{DBLP:journals/corr/abs-1802-00682}
M.~Narayanan, E.~Chen, J.~He, B.~Kim, S.~Gershman, and F.~Doshi{-}Velez.
\newblock How do humans understand explanations from machine learning systems?
  an evaluation of the human-interpretability of explanation.
\newblock {\em CoRR}, abs/1802.00682, 2018.

\bibitem{DBLP:conf/aaai/NarodytskaKRSW18}
N.~Narodytska, S.~Prasad Kasiviswanathan, L.~Ryzhyk, M.~Sagiv, and T.~Walsh.
\newblock Verifying properties of binarized deep neural networks.
\newblock In {\em Proc. of AAAI'18}, pages 6615--6624, 2018.

\bibitem{DBLP:conf/icml/ParmentierV21}
A.~Parmentier and T.~Vidal.
\newblock Optimal counterfactual explanations in tree ensembles.
\newblock In {\em Proceedings of the 38th International Conference on Machine
  Learning, {ICML} 2021, 18-24 July 2021, Virtual Event}, volume 139 of {\em
  Proceedings of Machine Learning Research}, pages 8422--8431. {PMLR}, 2021.

\bibitem{Lime16}
M.~T. Ribeiro, S.~Singh, and C.~Guestrin.
\newblock "why should {I} trust you?": Explaining the predictions of any
  classifier.
\newblock In {\em Proceedings of the 22nd {ACM} {SIGKDD} International
  Conference on Knowledge Discovery and Data Mining}, pages 1135--1144. {ACM},
  2016.

\bibitem{Anchor18}
M.~T. Ribeiro, S.~Singh, and C.~Guestrin.
\newblock Anchors: High-precision model-agnostic explanations.
\newblock In {\em Proc. of AAAI'18}, pages 1527--1535, 2018.

\bibitem{DBLP:journals/corr/abs-2103-11251}
C.~Rudin, C.~Chen, Z.~Chen, H.~Huang, L.~Semenova, and C.~Zhong.
\newblock Interpretable machine learning: Fundamental principles and 10 grand
  challenges.
\newblock {\em CoRR}, abs/2103.11251, 2021.

\bibitem{ShihChoiDarwiche18b}
A.~Shih, A.~Choi, and A.~Darwiche.
\newblock Formal verification of {B}ayesian network classifiers.
\newblock In {\em Proc. of PGM'18}, pages 427--438, 2018.

\bibitem{ShihCD18}
A.~Shih, A.~Choi, and A.~Darwiche.
\newblock A symbolic approach to explaining bayesian network classifiers.
\newblock In {\em Proceedings of the Twenty-Seventh International Joint
  Conference on Artificial Intelligence ({IJCAI}'18)}, pages 5103--5111, 2018.

\bibitem{ShihChoiDarwiche19}
A.~Shih, A.~Choi, and A.~Darwiche.
\newblock Compiling {B}ayesian networks into decision graphs.
\newblock In {\em Proc. of AAAI'19}, pages 7966--7974, 2019.

\bibitem{DBLP:conf/sat/ShihDC19}
A.~Shih, A.~Darwiche, and A.~Choi.
\newblock Verifying binarized neural networks by {A}ngluin-style learning.
\newblock In {\em Proc. of SAT'19}, pages 354--370, 2019.

\end{thebibliography}

\newpage
\section*{Proofs}

\noindent {\bf Proof of Proposition \ref{prop:complexityimplicanttestBT}}
\begin{proof}~\\
\begin{itemize}
\item Membership to {\sf coNP}: we consider the complementary problem and show that it belongs to {\sf NP}. In order to determine whether $t$ is 
not an abductive explanation for $\vec x$ given $BT$, it is enough to guess an instance $\vec x' \in \vec X$ such that $t \subseteq t_{\vec x'}$ 
and to check that $BT(\vec x') \neq BT(\vec x)$. Since the class associated by $BT$ to any input instance can be computed in time polynomial
in the size of $BT$ and the size of the instance, the conclusion follows.
\medskip
\item {\sf coNP}-hardness: it has been shown in \cite{Audemardetal22} (Proposition 3) that deciding whether $t$ is an abductive explanation for $\vec x$ given 
a random forest $RF$ over Boolean attributes is {\sf coNP}-complete. Thus, it is enough to show that we can associate in polynomial time any random
forest $RF = \{T_1, \ldots, T_p\}$ over Boolean attributes $A_1, \ldots, A_n$ to a boosted tree $BT = \{F\}$ with $F = \{T'_1, \ldots, T'_p\}$
such that for any $\vec x \in \vec X$, we have $RF(\vec x) = 1$ if and only if $BT(\vec x) = 1$. Each $T'_i$ ($i \in [p])$ is obtained in linear time from 
$T_i$ by replacing every $0$-leaf (resp. $1$-leaf) of $T_i$ by a leaf labelled by $-w$ (resp. $w$) where $w$ is a (fixed) positive number (e.g., $w = 0.5$).
By construction, we have $RF(\vec x) = 1$ if and only if $\sum_{j=1}^p T_i(\vec x) > \frac{p}{2}$ if and only if $\sum_{j=1}^p T'_i(\vec x) > 0$
if and only if $BT(\vec x) = 1$.
\end{itemize}
\end{proof}

\noindent {\bf Proof of Proposition \ref{prop:charact-worst}}
\begin{proof}
Suppose that $BT(\vec  x) = 1$, i.e., $w(F, \vec x) > 0$. By definition, $t$ is an abductive explanation for $\vec x$ given $BT$ if and only if any 
$\vec x' \in \vec X$ such that $t \subseteq t_{\vec x'}$ satisfies $BT(\vec x') = 1$. Since any $\vec x'' \in W(t, F)$ satisfies $t \subseteq t_{\vec x''}$,
we must have $BT(\vec x'') = 1$. Conversely, suppose that for any $\vec x'' \in W(t, F)$ we have $BT(\vec x'') = 1$. Then we have $w(F, \vec x'') > 0$.
By definition of $W(t, F)$, for any $\vec x' \in \vec X$ such that $t \subseteq t_{\vec x'}$, we have $w(F, \vec x') \geq w(F, \vec x'')$. 
Since $BT(\vec x'') = 1$, we have $w(F, \vec x'') > 0$, hence by transitivity of $>$, we get that $w(F, \vec x') >0$, or equivalently that
$BT(\vec x') = 1$.

Similarly, consider the case when $BT(\vec  x) = 0$, i.e., $w(F, \vec x) \leq 0$. By definition, $t$ is an abductive explanation for $\vec x$ given $BT$ if and only if any 
$\vec x' \in \vec X$ such that $t \subseteq t_{\vec x'}$ satisfies $BT(\vec x') = 0$. Since any $\vec x'' \in B(t, F)$ satisfies $t \subseteq t_{\vec x''}$,
we must have $BT(\vec x'') = 0$. Conversely, suppose that for any $\vec x'' \in B(t, F)$ we have $BT(\vec x'') = 0$. Then we have $w(F, \vec x'') \leq 0$.
By definition of $B(t, F)$, for any $\vec x' \in \vec X$ such that $t \subseteq t_{\vec x'}$, we have $w(F, \vec x') \leq w(F, \vec x'')$. 
Since $BT(\vec x'') = 0$, we have $w(F, \vec x'') \leq 0$, hence by transitivity of $\leq$, we get that $w(F, \vec x') \leq0$, or equivalently that
$BT(\vec x') = 0$.
\end{proof}

\noindent {\bf Proof of Proposition \ref{prop:charact-worst-multiclass}}
\begin{proof}
If $t$ is an abductive explanation for $\vec x$ given $BT$ , then for every $\vec x'$ extending $t$ we must have $BT(\vec x') = i$, that is 
$w(F^i, \vec x') > w(F^j, \vec x')$ for every $j \in [m] \setminus \{i\}$. This is equivalent to state that 
$w(F^i, \vec x') - \mathit{max}_{j \in [m] \setminus \{i\}} w(F^j, \vec x') > 0$. Since any worst instance $\vec x'$ extending $t$ given $BT$ and $\vec x$
is an instance that extends $t$, we have 
$$w(F^i, \vec x') - \mathit{max}_{j \in [m] \setminus \{i\}} w(F^j, \vec x') > 0,$$ as expected.

Conversely, suppose that for any worst instance $\vec x'$ extending $t$ given $BT$ and $\vec x$, we have 
$w(F^i, \vec x') - \mathit{max}_{j \in [m] \setminus \{i\}} w(F^j, \vec x') > 0$.  By definition, if $\vec x'$ is a worst instance 
extending $t$ given $BT$ and $\vec x$, then for any $\vec x'' \in \vec X$ that extends $t$, we have
$$w(F^i, \vec x') - \mathit{max}_{j \in [m] \setminus \{i\}} w(F^j, \vec x') \leq w(F^i, \vec x'') - \mathit{max}_{j \in [m] \setminus \{i\}} w(F^j, \vec x'').$$
Hence, if $w(F^i, \vec x') - \mathit{max}_{j \in [m] \setminus \{i\}} w(F^j, \vec x') > 0$, we also have that 
$w(F^i, \vec x'') - \mathit{max}_{j \in [m] \setminus \{i\}} w(F^j, \vec x'') > 0$, showing that $BT(\vec x'') = i$.
\end{proof}
    
\noindent {\bf Proof of Proposition \ref{prop:specific-abductive}}
\begin{proof}
Towards a contradiction, suppose that $BT(\vec x) = i \in [m]$ and there exists an instance $\vec x'$ extending $t$ and such that
$BT(\vec x') = j \in [m]$ with $j \neq i$. This implies that $w(F^j, \vec x') > w(F^k, \vec x')$ for every $k \in [m] \setminus \{j\}$.
Especially, for $k = i$, we have $w(F^j, \vec x') > w(F^i, \vec x')$.

Since $t$ is a tree-specific explanation for $\vec x$ given $BT$, 
$t$ is a subset of $t_{\vec x}$ such that for every $k \in [m] \setminus \{i\}$, 
we have $\sum_{l=1}^{p_i} w_\downarrow(t, T_l^i) > \sum_{l=1}^{p_k} w_\uparrow(t, T_l^k)$.
In particular, for $k = j$, we have $\sum_{l=1}^{p_i} w_\downarrow(t, T_l^i) > \sum_{l=1}^{p_j} w_\uparrow(t, T_l^j)$.

However, by definition of the utmost instances, for every $\vec x'$ extending $t$, we have $w(T_l^i, \vec x') \geq w_\downarrow(t, T_l^i)$
for every $T_l^i \in F^i$ and $w(T_l^k, \vec x') \leq w_\uparrow(t, T_l^k)$ for every $T_l^k \in F^k$ with $k \in [m] \setminus \{i\}$.
In particular, we have $w(T_l^j, \vec x') \leq w_\uparrow(t, T_l^j)$ for every $T_l^j \in F^j$.

Finally, we get that $w(F^j, \vec x') = \sum_{l=1}^{p_j} w(T_l^j, \vec x') \leq \sum_{l=1}^{p_j} w_\uparrow(t, T_l^j)
< \sum_{l=1}^{p_i} w_\downarrow(t, T_l^i) \leq \sum_{l=1}^{p_i} w(T_l^i, \vec x') = w(F^i, \vec x')$. A contradiction. 
\end{proof}

\noindent {\bf Proof of Proposition \ref{prop:discrepancy}}    
\begin{proof}
Consider $BT = \{F\}$ with $F = \{T_i^+, T_i^- : i \in [n]\}$ where for each $i \in [n]$, 
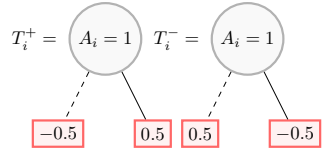
\begin{center}
\scalebox{0.7}{
      \begin{tikzpicture}[scale=0.9, roundnode/.style={circle, draw=gray!60, fill=gray!5, very thick, minimum size=7mm},
      squarednode/.style={rectangle, draw=red!60, fill=red!5, very thick, minimum size=5mm}]
        \node[roundnode](root) at (2,7){$A_i = 1$};
        \node at (0.5, 7){$T_i^+ =$};
        \node[squarednode](n1) at (1,5){$-0.5$};
        \node[squarednode](n2) at (3,5){$0.5$};
        \draw[dashed] (root) -- (n1);
        \draw(root) -- (n2);
        
        \node[roundnode](rootbis) at (5,7){$A_i = 1$};
        \node at (3.5, 7){$T_i^- =$};
        \node[squarednode](n1bis) at (4,5){$0.5$};
        \node[squarednode](n2bis) at (6,5){$-0.5$};
        \draw[dashed] (rootbis) -- (n1bis);
        \draw(rootbis) -- (n2bis);
        \end{tikzpicture}}
\end{center}

Consider the instance $\vec x = (0, \ldots, 0)$. We have $w(F, \vec x) = 0$, hence $F(\vec x) = 0$.
Consider any $i \in [n]$, let $\overline{(A_i=1)} \in t_{\vec x}$ and $t = t_{\vec x} \setminus \{\overline{A_i=1}\}$.
For every $j \in [n] \setminus \{i\}$, we have $B(t, T_j^+) = B(t, T_j^-) = \{\vec x\}$. We also have $B(t, T_i^-) = \{\vec x\}$.
Furthermore, $B(t, T_i^+) = \{\vec x'\}$ where $\vec x'$ is the instance that coincides with $\vec x$, except that $(A_i=1) \in t_{\vec x'}$.
Accordingly,  $\sum_{j=1}^n (w_\uparrow(t, T_j^+) + w_\uparrow(t, T_j^-)) = 1 > 0$, showing that $t$  is not a tree-specific 
explanation for $\vec x$ given $F$. Since the weights of the utmost instances extending a term $t$ given $BT = \{F\}$ 
varies monotonically when $t$ is deprived of some of its elements and since $\sum_{j=1}^n (w_\uparrow(t_{\vec x}, T_j^+) + w_\uparrow(t_{\vec x}, T_j^-)) = 0$,
we can conclude that $t_{\vec x}$ is the unique tree-specific explanation for $\vec x$ given $F$. Contrastingly, since for every $\vec x' \in \vec X$,
we have $F(\vec x') = 0$, $\emptyset$ is the (unique) sufficient reason for $\vec x$ given $F$.
\end{proof}

\noindent {\bf Proof of Proposition \ref{prop:correctness}}
%
\begin{proof}
The proof consists of two points. First, we check that for every $j \in [m] \setminus \{i\}$ (where $BT(\vec x) = i$), we have 
$\sum_{k=1}^{p_i} w_\downarrow(t_{\vec x}, T_k^i) > \sum_{k=1}^{p_j} w_\uparrow(t_{\vec x}, T_k^j)$ holds.
Since $\vec x$ is the unique instance that extends $t_{\vec x}$, for any tree $T_k^l$ of $BT$, $\vec x$ is also a worst and a best instance extending $t_{\vec x}$ given $T_k^l$. Thus, for each $k \in [p_i]$, we have $w_\downarrow(t_{\vec x}, T_k^i) = w(T_k^i, \vec x)$ and for each $k \in [p_j]$, we have $w_\uparrow(t_{\vec x}, T_k^j) = w(T_k^j, \vec x)$. Acccordingly, $\sum_{k=1}^{p_i} w_\downarrow(t_{\vec x}, T_k^i) > \sum_{k=1}^{p_j} w_\uparrow(t_{\vec x}, T_k^j)$  is equivalent to
$\sum_{k=1}^{p_i} w(T_k^i, \vec x) > \sum_{k=1}^{p_j} w(T_k^j, \vec x)$, which is equivalent to $w(F^i, \vec x) > w(F^j, \vec x)$ and finally to $BT(\vec x) = i$,
which holds.

The second point consists in verifying that if $t, t'$ verify $t \subset t' \subseteq t_{\vec x}$, and for every $j \in [m] \setminus \{i\}$, we have
$\sum_{k=1}^{p_i} w_\downarrow(t', T_k^i) \leq \sum_{k=1}^{p_j} w_\uparrow(t', T_k^j)$ holds, then
$\sum_{k=1}^{p_i} w_\downarrow(t, T_k^i) \leq \sum_{k=1}^{p_j} w_\uparrow(t, T_k^j)$ holds as well.
This comes directly from the fact that when $t \subset t'$, we have $w_\downarrow(t, T_k^i) \leq w_\downarrow(t', T_k^i)$ for each $k \in [p_i]$
and we have $w_\uparrow(t, T_k^j) \geq w_\uparrow(t', T_k^j)$ for each $k \in [p_j]$.
\end{proof}

\end{document}